\documentclass{article} %
\usepackage{iclr2027_conference,times}

\usepackage{amsmath,amsfonts,bm}

\def\eqref#1{equation~\ref{#1}}

\def\1{\bm{1}}

\DeclareMathAlphabet{\mathsfit}{\encodingdefault}{\sfdefault}{m}{sl}
\SetMathAlphabet{\mathsfit}{bold}{\encodingdefault}{\sfdefault}{bx}{n}

\usepackage{hyperref}
\usepackage{url}

\usepackage{placeins}
\usepackage{float}
\usepackage[utf8]{inputenc}
\usepackage[T1]{fontenc}
\usepackage{booktabs}
\usepackage{amsfonts,amsmath,amssymb}
\usepackage{nicefrac}
\usepackage{graphicx}
\usepackage{microtype}
\usepackage{xcolor}
\usepackage{subcaption}
\usepackage{multirow}
\usepackage{array}
\usepackage{cleveref}
\usepackage{bm}
\usepackage{mathtools}
\usepackage{enumitem}
\usepackage{tikz}
\usepackage{stfloats}
\usepackage{tabularx}
\usepackage{multirow}
\usepackage{booktabs}
\usetikzlibrary{shapes.geometric, arrows.meta, positioning, fit, backgrounds, calc, decorations.pathreplacing}

\definecolor{frozenblue}{RGB}{74,123,183}
\definecolor{headpurple}{RGB}{117,107,177}
\definecolor{trainonlyteal}{RGB}{0,158,115}
\definecolor{contribink}{RGB}{86,70,143}
\definecolor{neutralgray}{RGB}{245,245,245}
\definecolor{lossorange}{RGB}{230,159,0}
\definecolor{centeringcolor}{RGB}{0,158,115}
\definecolor{bridgecolor}{RGB}{117,107,177}

\DeclareMathOperator{\sg}{sg}

\title{Can a world model predict a deepfake video? A Consistency Margin Prior for Deepfake Detection}

\author{%
\makebox[\textwidth][c]{%
\shortstack[c]{%
\textbf{Pradip Gupta, Shubham Mandloi, Ravi Kumar Satzoda,}\\[5pt]
\textbf{Sujeeth Bharadwaj, Yassine Benajiba, Dan Roth}\\[8pt]
\normalfont Oracle AI
}%
}%
}

\iclrfinalcopy %
\begin{document}

\maketitle

\begin{abstract}
Deepfake detectors often perform well on the dataset they are trained on,
but their performance drops when tested on videos from different datasets.
In this paper, we study whether a predictive video model can provide a
signal that generalizes better across datasets. Instead of relying only on
generator-specific artifacts, as many existing detectors do, we use a model
trained to predict natural video and measure the difference between what it
predicts and what it observes in a video.

Using V-JEPA~2.1, we keep the encoder frozen and adapt the predictor using
rank-16 LoRA. We define the difference between the predictor output and the
frozen encoder target as the \emph{prediction residual}. We hypothesize that
manipulated clips produce, on average, larger prediction residuals than
authentic clips, and use this as our consistency-margin prior. We process
these residuals using a lightweight Latent Consistency Head (LCH), which
separates residual magnitude from direction for training and classification.

To test whether the direction of the consistency margin affects cross-dataset generalization, we compare standard and reversed margin objectives while keeping the remaining training setup fixed. We match the two settings on FF++ validation performance before evaluating them on unseen data. The reversed-margin setting exhibits a larger cross dataset generalization gap on average, providing evidence that the direction of the consistency margin affects cross-dataset transfer.
Our complete model adds only $8.25$\,M trainable parameters, which is
$2.47\%$ of the total $334$\,M parameters. Trained only on FF++, it
achieves an AUC of $90.5$ on Celeb-DFv2 and performs well across several
other cross-dataset benchmarks. Our ablations show that separating residual
magnitude from direction, reducing dataset-specific variation, and using a
low-rank projection improve cross-dataset generalization in our experiments.
\end{abstract}

\section{Introduction}
\label{sec:intro}

Deepfake detectors often achieve strong performance on the datasets they are trained on, but their performance drops when they are tested on data from different sources. XceptionNet~\citep{chollet2017xception,rossler2019faceforensics} and
SBI~\citep{shiohara2022sbi}, for example, achieve strong results on FF++, while
cross-dataset evaluation on Celeb-DFv2~\citep{li2020celebdf},
DFDC~\citep{dolhansky2020dfdc}, and WildDeepfake~\citep{zi2020wilddeepfake}
remains much more difficult. Recent benchmarks show the same
pattern~\citep{chandra2025deepfakeeval,df402024,gend2025,ntire2026deepfake}.
One reason is that detectors learn dataset-specific cues, such as compression
traces, generator fingerprints, or collection biases. These cues may work well on the
training dataset but may not generalize to new datasets
~\citep{cao2025frequency,wang2023altfreezing,guo2025tfcu,yan2023ucf,hu2025plada}.
This motivates us to look for a signal that is less dependent on a specific dataset
or deepfake generator.
\label{sec:intro:crisis}

\begin{figure*}[t]
    \centering
    \includegraphics[
        width=\textwidth
    ]{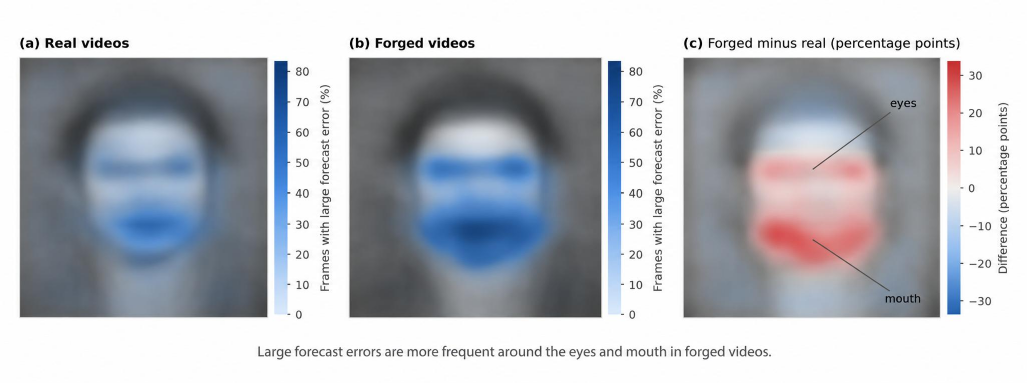}
    \caption{\textbf{Spatial distribution of prediction errors.}
    Average spatial maps of large forecast errors for real and forged
    videos, together with their difference.}
    \label{fig:forecast_error_heatmaps}
\end{figure*}

We explore this idea using the self-supervised video prediction capabilities of
V-JEPA~2.1~\citep{vjepa2_1_2025}. We keep the V-JEPA encoder frozen and adapt its predictor using rank-16 LoRA on all 12 predictor blocks. We then compare the predictor output with the corresponding stop-gradient encoder target. We call the difference between them the \textit{prediction residual} ($r$). Our main hypothesis is that manipulated clips should, on average, produce larger prediction residuals than authentic clips (Fig. \ref{fig:forecast_error_heatmaps}). We define this as the consistency-margin prior, i.e., $\|r_{\rm fake}\| > \|r_{\rm real}\|$.

To use these residuals for deepfake detection, we introduce the \textit{Latent Consistency Head (LCH)}, a lightweight detector built on top of the V-JEPA~2.1 predictor. LCH separates the residual into magnitude and
direction. The residual magnitude is used by the consistency loss, while the
residual direction is used to classify the video as real or fake. Additionally, we will show how LCH processes the residual using our Predictor Attention-conditioned Stop-
gradient Centering (PASC), horizon-specific projections, and spatial anomaly pooling to learn features for cross-dataset detection. We show that although trained only on FF++, the detector performs well across several cross-dataset benchmarks (Celeb-DFv2, DFDC, WildDeepfake, and Deepfake-Eval-2024), indicating that prediction residuals can provide a useful signal across datasets.

We further examine whether the direction of the consistency margin affects cross-dataset generalization. We introduce the Sign-Flip Identification Protocol (SFIP), which compares a standard setting that encourages larger residual magnitudes for manipulated clips with a reversed setting that encourages larger residual magnitudes for authentic clips. The two settings
keep the remaining training setup fixed and are matched on FF++ validation performance before cross-dataset evaluation. Across the evaluated cross-dataset benchmarks, the reversed-margin setting has an average generalization gap of 1.8\% larger than the standard setting, indicating worse cross-dataset transfer under the reversed margin. We describe SFIP in Section~\ref{sec:method:sfip}. Furthermore, Our model remains parameter-efficient, adding only a small set of trainable parameters to V-JEPA~2.1. The LCH contains $6.31$M trainable parameters, while predictor adaptation contributes an additional $1.48$M.

\textbf{Contributions.}
\label{sec:intro:contributions}
\begin{enumerate}[leftmargin=*, nosep]
    \item \textbf{A latent prediction-residual formulation for deepfake detection.} We introduce a video-level forensic formulation based on V-JEPA~2.1 prediction residuals, extending residual-based deepfake detection from image reconstruction to self-supervised latent video prediction.

    \item \textbf{Latent Consistency Head.} We propose LCH, a lightweight detector that separates residual magnitude from direction and combines PASC, horizon-specific projections, and spatial anomaly pooling to learn features for cross-dataset real/fake classification.

    \item \textbf{A matched sign-flip test for the consistency margin prior.} We introduce SFIP to compare standard and reversed consistency-margin objectives at matched FF++ validation performance. Across the evaluated cross-dataset benchmarks, reversing the margin increases the average generalization gap by 1.8\%, providing evidence that margin direction affects cross-dataset transfer.
\end{enumerate}

\section{Related Work}
\label{sec:related}

\textbf{Frozen foundation models for forensics.}
Recent deepfake detectors adapt large pretrained models instead of training
the full backbone. Forensics Adapter~\citep{cui2025forensicsadapter},
LNCLIP-DF~\citep{tan2025lnclip}, DeepFake-Adapter~\citep{shao2024deepfakeadapter},
and UniFD~\citep{ojha2023towards} show that frozen or lightly adapted
foundation models can transfer well across datasets but current works do not use predictive video models like V-JEPA and their prediction results for deepfake detection".

\textbf{Reconstruction and residual signals.}
Several deepfake detectors use reconstruction error or consistency as a
forensic signal~\citep{zhao2021selfconsistency,khalid2020ocfakedet,
wang2022cscl,cao2022recce,dong2023caddm,wang2025fsfm}.
DIRE~\citep{wang2023dire}, for example, detects generated images using
diffusion reconstruction errors, while NPR~\citep{tan2024npr} uses
neighboring-pixel relationships. Our method follows the same general idea
of using prediction error, but computes it in the latent space of a
self-supervised video predictor.

\textbf{Cross-dataset deepfake detection.}
Many methods improve cross-dataset detection through augmentation,
temporal modeling, feature disentanglement, or more general
representations~\citep{nguyen2024laanet,shiohara2022sbi,yan2023ucf,
zheng2021ftcn,wang2023altfreezing,guo2025tfcu,haliassos2021lipforensics}.
Other methods use frequency information to capture forgery
artifacts~\citep{qian2020f3net,tan2024freqnet,baru2025waveletclip,
cao2025frequency}. However, these cues can still depend on the generator,
compression pipeline, or dataset. We instead study whether prediction
residuals from a self-supervised video model provide a signal that
generalizes across datasets.

\textbf{Relationship to existing components.}
The individual operations used in our detector are related to existing
ideas. The Bridge uses low-rank projection~\citep{lora2022hu}, residual
normalization is related to normalized embedding methods
~\citep{wang2017normface}, PASC is related to grouped normalization
~\citep{wu2018groupnorm}, and our consistency objective uses a margin
loss~\citep{schroff2015facenet}. Our contribution is not these operations
individually, but how they are combined to classify predictive-video
residuals and how we test the consistency-margin direction using SFIP.
Additional comparisons with recent detectors are provided in
Appendix~\ref{app:extended_related_work}.

\section{Method}
\label{sec:method}
In this section, we now describe the proposed architecture - ConsistencyNet (Fig.~\ref{fig:consistency_net}) that utlizes the prediction residual introduced in Section~\ref{sec:intro}.

\subsection{Problem Formulation}
\label{sec:method:problem}

\paragraph{Prediction residual.}
For a video clip $\mathcal{V}$, we form two inputs from the same
16 frames and pass them through the V-JEPA~2.1 encoder. The first
contains the complete, non-masked frames, while the second applies a
spatio-temporal mask to the same frames. The encoder produces a
temporal feature representation for each input, i.e., $
\mathbf{H}_{\rm full},\,
\mathbf{H}_{\rm mask}
\in
\mathbb{R}^{T\times S\times1024}$ where $T=8$ is the number of temporal feature positions and $S=576$ is the
number of spatial tokens at each temporal position. We omit the batch
dimension for clarity. We use two prediction lengths,
$\delta\in\mathcal{D}=\{2,4\}$. For a given $\delta$, the first
$T-\delta$ temporal positions of the masked representation are used
as context resulting in $\mathbf{h}_{\rm ctx}^{(\delta)}$. The V-JEPA~2.1 predictor $P_\psi$ uses these context features to predict the representations of the final $\delta$ temporal positions - $\mathbf{h}_{P}^{(\delta)}$. Finally, for the same final $\delta$ temporal positions, we take the corresponding features from the non-masked encoder representation to generate $\mathbf{h}_{E}^{(\delta)}$. These are represented in Eq. 1 - \ref{eq:encoder-target} below.
\begin{align}
\mathbf{h}_{\mathrm{ctx}}^{(\delta)}
&= \mathbf{H}_{\mathrm{mask}}[0 : T - \delta]
\in \mathbb{R}^{(T-\delta)S \times 1024},
\label{eq:context-features} \\[0.5em]
\mathbf{h}_{P}^{(\delta)}
&= P_{\psi}\!\left(\mathbf{h}_{\mathrm{ctx}}^{(\delta)}\right)
\in \mathbb{R}^{\delta S \times 1664},
\label{eq:predictor-output} \\[0.5em]
\mathbf{h}_{E}^{(\delta)}
&= \mathbf{H}_{\mathrm{full}}[T-\delta : T]
\in \mathbb{R}^{\delta S \times 1024}.
\label{eq:encoder-target}
\end{align}

Thus, $\mathbf{h}_{P}^{(\delta)}$ is the representation predicted
from the masked context, while $\mathbf{h}_{E}^{(\delta)}$ is the
representation of the same temporal positions obtained from the
non-masked encoder input. The predictor output is 1664-dimensional, while the corresponding
encoder representation is 1024-dimensional. We therefore use a
low-rank Bridge $\mathcal{B} = B_2\!\left(B_1(\mathbf{h})\right)$ consisting of two linear projections, $B_1=\operatorname{Linear}(1664,64)$ and $B_2=\operatorname{Linear}(64,1024)$ to map the predictor output to the encoder feature space. The Bridge contains 0.17\,M trainable parameters. The prediction residual
is then computed as defined in Eq.~\ref{eq:residual}.
\begin{equation}
\mathbf{r}^{(\delta)}
=
\mathcal{B}\!\left(\mathbf{h}_{P}^{(\delta)}\right)
-
\sg\!\left(\mathbf{h}_{E}^{(\delta)}\right)
\in
\mathbb{R}^{\delta S\times1024},
\label{eq:residual}
\end{equation}
where $\sg(\cdot)$ denotes stop-gradient. The residual therefore
measures the difference between what the predictor estimates from the
masked context and what the encoder represents at the corresponding
positions in the complete video. A qualitative example of prediction residual is provided in Appendix~\ref{app:residual_visualization}

\subsection{Architecture Overview}
\label{sec:method:overview}

\begin{figure*}[!t]
    \centering
    \includegraphics[
        width=\textwidth,
        clip
    ]{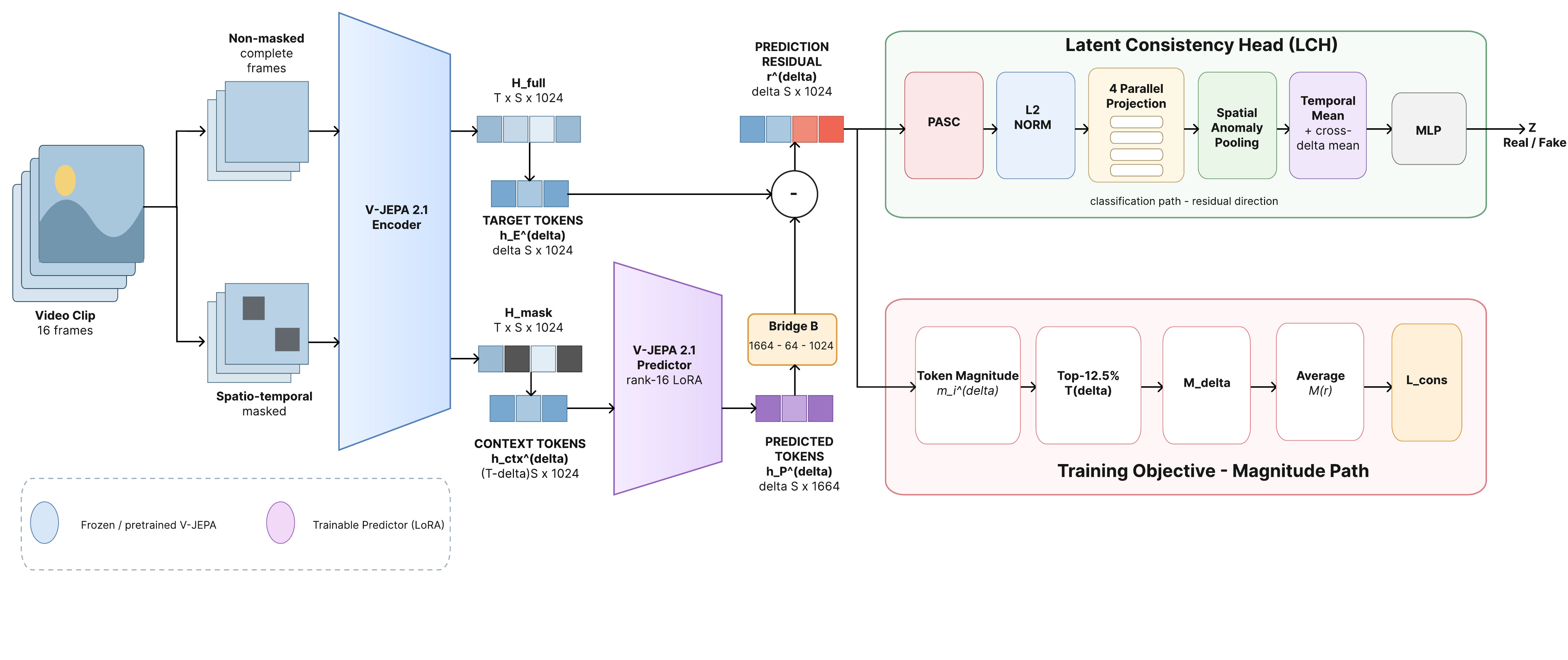}
    \caption{\textbf{Overview of ConsistencyNet.} V-JEPA~2.1 constructs prediction residuals for prediction lengths $\delta\in\{2,4\}$. LCH processes the residual using PASC, $\ell_2$ normalization, parallel projections, SAP, and temporal aggregation before real/fake classification. Residual magnitude is used separately by the consistency-margin objective.}
    \label{fig:consistency_net}
\end{figure*}

Fig.~\ref{fig:consistency_net} shows the overall architecture of
ConsistencyNet. We use the V-JEPA~2.1 encoder and predictor to
construct the prediction residual defined in
Section~\ref{sec:method:problem}. The encoder has 304\,M parameters,
24 transformer blocks, and an output dimension of 1024. The predictor
has 21.9\,M base parameters and 12 transformer blocks. The prediction residual $\mathbf{r}^{(\delta)}$ is the input to our
\textit{Latent Consistency Head (LCH)}, which produces the real/fake
prediction. For each prediction length
$\delta\in\mathcal{D}=\{2,4\}$, LCH applies Predictor Attention-conditioned Stop-gradient Centering (PASC) and $\ell_2$
normalization, followed by four parallel projection heads and Spatial
Anomaly Pooling (SAP). The resulting features are aggregated over the
predicted temporal positions and then across the two prediction
lengths before being passed to the final classifier. The raw residual
magnitude is handled separately by the training objective described in
Section~\ref{sec:method:training}.

\paragraph{Predictor Attention-conditioned Stop-gradient Centering (PASC).}
The predictor may rely on different amounts of visible context when
predicting different target tokens. Centering all residual tokens with
a single mean can therefore mix tokens that were predicted using
different amounts of context information. PASC addresses this by
grouping target tokens according to their attention to the visible
context and centering the two groups separately.

We use attention from block~6 of the predictor. Let
$A_{i,j}^{(6)}$ denote the attention from target token $i$ to token
$j$, averaged over attention heads. We define the context-attention
ratio for target token $i$ as $c_i = \sum_{j\in\mathrm{ctx}} A_{i,j}^{(6)}/\sum_j A_{i,j}^{(6)}$. Here, $\mathrm{ctx}$ denotes the visible context tokens. A larger
$c_i$ indicates that the prediction for token $i$ relies more strongly
on the visible context.

For each clip, we split the target tokens at the median
context-attention ratio as $\mathcal{H} = \{i:c_i>\operatorname{median}(c)\}$ and $\mathcal{L} = \{i:c_i\leq\operatorname{median}(c)\}$, where $\mathcal{H}$ and $\mathcal{L}$ denote the high- and low-context
groups. Each residual token is then centered using the mean residual
of its corresponding group, i.e., 
\begin{equation}
\tilde{\mathbf{r}}_i^{(\delta)}
=
\mathbf{r}_i^{(\delta)}
-
\frac{1}{|\mathcal{G}(i)|}
\sum_{j\in\mathcal{G}(i)}
\mathbf{r}_j^{(\delta)},
\qquad
\mathcal{G}(i)\in\{\mathcal{H},\mathcal{L}\}.
\label{eq:pasc}
\end{equation}
The grouping is label-free, and the attention used to define the
groups is detached from the gradient; gradients still flow through
the centered residuals. PASC adds no trainable parameters. At
inference, the attention hook is removed and group centering is
replaced by a single per-frame spatial mean.

\paragraph{Residual direction.}
After PASC, we normalize each centered residual token, i.e., $ \hat{\mathbf{r}}_i^{(\delta)} =\tilde{\mathbf{r}}_i^{(\delta)}/
    \max\!\left(
        \|\tilde{\mathbf{r}}_i^{(\delta)}\|_2,
        \sqrt{\varepsilon}
    \right)$, where $\varepsilon=10^{-12}$. This removes the raw residual magnitude from the classification path,
so the remaining LCH components operate on the normalized residual
direction.

\paragraph{Prediction-length-specific projections.}
For each prediction length $\delta\in\mathcal{D}$, we use $K=4$
parallel heads to project the normalized residual from 1024 to 768
dimensions,

\begin{equation}
\mathbf{e}_{i}^{(\delta,k)}
=
\operatorname{LN}\!\left(
W_{\delta}^{(k)}\hat{\mathbf{r}}_i^{(\delta)}
+\mathbf{b}_{\delta}^{(k)}
\right),
\quad
W_{\delta}^{(k)}\in\mathbb{R}^{768\times1024},
\quad k=1,\ldots,K.
\label{eq:parallel-projection}
\end{equation}

A learned softmax gate combines the four outputs,

\begin{equation}
\gamma_{\delta,1:K}
=
\operatorname{softmax}\!\left(
g_{\rm head}(\mathbf{e}^{(\delta,1:K)})
\right),
\qquad
\mathbf{e}_i^{(\delta)}
=
\sum_{k=1}^{K}
\gamma_{\delta,k}\mathbf{e}_i^{(\delta,k)}
\in\mathbb{R}^{768}.
\label{eq:projection-gate}
\end{equation}

\paragraph{Spatial Anomaly Pooling (SAP).}
For prediction length $\delta$, each of the $\delta$ predicted
temporal positions contains $S=576$ projected spatial-token features.
SAP normalizes these spatial features within each sample and uses a
learned softmax gate to give more weight to informative spatial
locations. It reduces the spatial dimension as $[B,\delta S,768] \longrightarrow [B,\delta,768]$. We denote the pooled representation at predicted temporal position
$t$ by $\mathbf{p}_t^{(\delta)}$.

\paragraph{Temporal and cross-prediction-length aggregation.}
For each prediction length, we first average the SAP features over its
$\delta$ predicted temporal positions. We then average the resulting
representations across the two prediction lengths,

\begin{equation}
\begin{aligned}
\mathbf{F}_{\delta}
&=
\frac{1}{\delta}
\sum_{t=1}^{\delta}
\mathbf{p}_t^{(\delta)}
\in\mathbb{R}^{768},
\qquad
&
\mathbf{F}
&=
\frac{1}{|\mathcal{D}|}
\sum_{\delta\in\mathcal{D}}
\mathbf{F}_{\delta},
\qquad
\mathcal{D}=\{2,4\}.
\end{aligned}
\label{eq:temporal-cross-aggregation}
\end{equation}

\paragraph{Classifier.}
Finally, a two-layer MLP ($768\rightarrow384\rightarrow2$) maps the
combined representation $\mathbf{F}$ to the real/fake logits
$\mathbf{z}=\operatorname{MLP}(\mathbf{F})$. Detailed implementation information is provided in Appendix~\ref{app:method_details}.

\subsection{Training Objective}
\label{sec:method:training}

We train ConsistencyNet using two losses. The consistency-margin loss
acts on the raw prediction residual, while the classification loss
supervises the real/fake prediction from LCH.

\paragraph{Consistency-margin prior.}
As introduced in Section~\ref{sec:intro}, we hypothesize that manipulated
clips produce larger prediction residuals than authentic clips on average,
$\mathbb{E}[\|\mathbf{r}_{\rm fake}\|_2] >
\mathbb{E}[\|\mathbf{r}_{\rm real}\|_2]$.

Since V-JEPA is trained to predict natural-video representations,
manipulations may increase the mismatch between predicted and observed
features. We therefore treat it as an
empirical training prior, not a property guaranteed for every clip.
Further explanation is provided in Appendix~\ref{app:jepa_to_prior}.

\paragraph{Residual magnitude score.}
To use this prior during training, we compute the raw magnitude of each
residual token as
$m_i^{(\delta)}=\|\mathbf{r}_i^{(\delta)}\|_2$, where $i$ indexes a
residual token for prediction length $\delta$.

Prediction errors may be concentrated in only a subset of spatial
locations. For each prediction length $\delta$, we therefore select
the largest $12.5\%$ of the token magnitudes and average them,

\begin{equation}
M_{\delta}\!\left(\mathbf{r}^{(\delta)}\right)
=
\frac{1}{|\mathcal{T}_{\delta}|}
\sum_{i\in\mathcal{T}_{\delta}}
m_i^{(\delta)},
\qquad
\mathcal{T}_{\delta}
=
\operatorname{TopK}_{12.5\%}
\left(
\{m_i^{(\delta)}\}
\right).
\label{eq:topk-magnitude-offset}
\end{equation}

We then average the magnitude scores across the two prediction
lengths,

\begin{equation}
M(\mathbf{r})
=
\frac{1}{|\mathcal{D}|}
\sum_{\delta\in\mathcal{D}}
M_{\delta}\!\left(\mathbf{r}^{(\delta)}\right),
\qquad
\mathcal{D}=\{2,4\}.
\label{eq:topk-magnitude}
\end{equation}

A larger $M(\mathbf{r})$ indicates a stronger prediction mismatch in
the locations with the largest residuals.

\paragraph{Consistency-margin loss.}
Using the magnitude score above, we encourage manipulated clips to
have a larger residual magnitude than authentic clips,

\begin{equation}
\mathcal{L}_{\rm cons}
=
\max\!\left(
0,\,
m
+
M(\mathbf{r}_{\rm real})
-
M(\mathbf{r}_{\rm fake})
\right),
\qquad
m=2.0.
\label{eq:consistency-margin-loss}
\end{equation}

The loss is active when the residual magnitude of a manipulated clip
is not larger than that of an authentic clip by at least the margin
$m$.

\paragraph{Classification loss.}
We train the LCH output using weighted cross-entropy,
$\mathcal{L}_{\rm wCE}
=-\frac{1}{N}\sum_{n=1}^{N}w_{y_n}
\log p(y_n\mid\mathcal{V}_n)$,
to account for the class imbalance in FF++, where $y_n$ is the
ground-truth label and $w_{y_n}$ is its class weight.

\paragraph{Total objective.}
The final training objective combines the classification and
consistency losses,

\begin{equation}
\mathcal{L}
=
\mathcal{L}_{\rm wCE}
+
\lambda\mathcal{L}_{\rm cons},
\qquad
\lambda=0.1.
\label{eq:total-loss}
\end{equation}

Thus, $\mathcal{L}_{\rm wCE}$ supervises the real/fake classifier,
while $\mathcal{L}_{\rm cons}$ acts on the raw prediction residual.

\subsection{Sign-Flip Identification Protocol (SFIP)}
\label{sec:method:sfip}

We use the Sign-Flip Identification Protocol (SFIP) to test whether
the direction of the consistency margin affects cross-dataset
generalization while keeping the remaining training setup fixed.

\paragraph{Standard and reversed objectives.}
The standard objective encourages manipulated clips to have a larger
residual magnitude than authentic clips,
$M(\mathbf{r}_{\rm fake}) > M(\mathbf{r}_{\rm real}) + m$.
The reversed objective encourages the opposite relation,
$M(\mathbf{r}_{\rm real}) > M(\mathbf{r}_{\rm fake}) + m$.

\paragraph{Matched in-domain performance.}
To control for in-domain differences, we match the standard and reversed
settings on FF++ validation AUC before cross-dataset evaluation, yielding
models with similar in-domain performance but opposite margin directions.

\paragraph{Cross-dataset comparison.}
For an out-of-domain dataset, we define the generalization gap as
\begin{equation}
G = \mathrm{AUC}_{\rm FF++} - \mathrm{AUC}_{\rm OOD},
\label{eq:generalization-gap}
\end{equation}
where $\mathrm{AUC}_{\rm OOD}$ is the AUC on the unseen dataset.
We compare the standard and reversed settings using
\begin{equation}
\Delta G = G_{\rm reverse} - G_{\rm standard}.
\label{eq:sfip-gap-difference}
\end{equation}
A positive $\Delta G$ indicates that the reversed-margin model has a
larger cross-dataset generalization gap at matched in-domain
performance.

\section{Experiments and Analysis}
\label{sec:experiments}

\subsection{Training and Evaluation}
\label{sec:exp:setup}

\paragraph{Cross-dataset evaluation.}
We train ConsistencyNet on FaceForensics++ using 16-frames and evaluate the cross-dataset generalization of our model on five unseen datasets. ConsistencyNet achieves 90.5 AUC on Celeb-DF-v2, 87.1 on DFDC,
90.4 on DFDCP, and 88.2 on WildDeepfake. While individual state-of-the-art
methods obtain higher AUC on specific benchmarks, their performance varies
considerably across datasets. In contrast, ConsistencyNet maintains strong
performance across all four established cross-dataset benchmarks, with AUC
consistently above 87. On the more recent DE-2024 benchmark, ConsistencyNet
achieves 66.4 AUC, outperforming the evaluated Effort and FSFM
checkpoints on the same test split. These results suggest that the proposed
prediction-residual representation provides consistent transfer across
different manipulation sources and evaluation domains. Full training, preprocessing,
and hyperparameter details are provided in Appendix~\ref{app:hparams}. 

\begin{table}[!htbp]
\centering
\scriptsize
\caption{\textbf{Cross-dataset generalization.}
Video-level AUC (\%) on unseen datasets. All methods are trained on FF++.
For DE-2024$^{*}$, values unavailable in the original papers are evaluated
by us using the authors' released checkpoints and inference pipelines.}
\label{tab:crossdataset}

\setlength{\tabcolsep}{3.4pt}
\renewcommand{\arraystretch}{1.03}

\resizebox{\textwidth}{!}{%
\begin{tabular}{@{}llccccc@{}}
\toprule
\textbf{Method}
& \textbf{Venue}
& \textbf{CDF-v2}
& \textbf{DFDC}
& \textbf{DFDCP}
& \textbf{WildDF}
& \textbf{DE-2024}$^{*}$ \\
\midrule

\multicolumn{7}{l}{\textit{Frame-based}} \\

CDFA~\citep{lin2024cdfa}
& ECCV'24 & \underline{97.2} & 83.8 & \textbf{97.0} & 84.5 & --- \\

Real appearance modelling~\citep{tian2024ram}
& ECCV'24 & 95.5 & 80.2 & 89.5 & 88.4 & --- \\

RepDFD~\citep{lin2025repdfd}
& AAAI'25 & 89.9 & 81.0 & \underline{95.0} & 88.1 & --- \\

FSFM~\citep{wang2025fsfm}
& CVPR'25 & 91.4 & 83.5 & 89.7 & 87.0 & 65.4 \\

Effort~\citep{yan2025effort}
& ICML'25 & 95.6 & 84.3 & 90.9 & 84.8 & 65.7 \\

Forensics Adapter~\citep{cui2025forensicsadapter}
& CVPR'25 & 95.7 & 87.2 & 92.9 & --- & 62.3 \\

DFF-Adapter~\citep{zhang2026dffadapter}
& AAAI'26 & 95.3 & \textbf{90.0} & 91.6 & --- & --- \\

DFD-HR~\citep{sun2026dfdhr}
& CVPR'26 & 96.0 & 86.5 & 90.1 & \underline{90.7} & --- \\

DiffusionFF~\citep{peng2026diffusionff}
& CVPR'26 & \textbf{97.2} & 85.1 & 92.6 & --- & --- \\

Query-token forgery purification~\citep{wang2026qtfp}
& CVPR'26 & 96.0 & 86.9 & 92.5 & \textbf{91.7} & --- \\

TSRL~\citep{lei2026tsrl}
& CVPR'26 & 90.1 & \underline{88.2} & 92.4 & --- & --- \\

\midrule
\multicolumn{7}{l}{\textit{Video-based}} \\

NACO~\citep{zhang2024naco}
& ECCV'24 & 89.5 & 76.7 & --- & --- & --- \\

Pixel-wise temporal frequency~\citep{kim2025temporalfreq}
& ICCV'25 & 89.7 & 75.2 & --- & --- & --- \\

DFD-FCG~\citep{han2025dfdfcg}
& CVPR'25 & 95.0 & 81.8 & --- & 87.2 & --- \\

TFCU~\citep{guo2025tfcu}
& CVPR'25 & 93.2 & 86.1 & --- & --- & --- \\

Video Blending + StA~\citep{yan2025vbsta}
& CVPR'25 & 94.7 & 84.3 & 90.9 & 84.8 & --- \\

FakeRadar~\citep{li2025fakeradar}
& ICCV'25 & 91.7 & 84.1 & 88.5 & --- & --- \\

FakeSTormer~\citep{nguyen2025fakestormer}
& ICCV'25 & 92.8 & 75.1 & 90.2 & 75.3 & 57.1 \\

\midrule
\textbf{ConsistencyNet (ours)}
& ---
& 90.5
& 87.1
& 90.4
& 88.2
& \textbf{66.4} \\

\bottomrule
\end{tabular}%
}

\vspace{2pt}
\begin{minipage}{\textwidth}
\scriptsize
$^{*}$For DE-2024, results not reported by the corresponding papers are
evaluated by us using the authors' released checkpoints and inference
pipelines. ConsistencyNet is evaluated on the official DE-2024 test split.
\end{minipage}

\end{table}

\FloatBarrier

{
\begin{table}[!htbp]
\centering
\scriptsize
\caption{\textbf{Cross-manipulation evaluation on DF40 face swap.}
Video-level AUC (\%) on eight unseen face-swap generators.}
\label{tab:df40swap}

\setlength{\tabcolsep}{2.5pt}
\renewcommand{\arraystretch}{1.04}

\resizebox{\textwidth}{!}{%
\begin{tabular}{@{}llccccccccc@{}}
\toprule
\textbf{Method}
& \textbf{Venue}
& \textbf{Blend}
& \textbf{e4s}
& \textbf{FD}
& \textbf{FSGAN}
& \textbf{IS}
& \textbf{MS}
& \textbf{SS}
& \textbf{UF}
& \textbf{Avg.} \\
\midrule

RECCE~\citep{cao2022recce}
& CVPR'22
& 83.2 & 68.3 & 84.8 & 94.9 & 84.8 & 92.5 & 76.8 & 89.8 & 84.4 \\

SLADD~\citep{chen2022sladd}
& CVPR'22
& 88.2 & 76.5 & 82.5 & 94.3 & 87.9 & 95.4 & 79.4 & 87.8 & 86.5 \\

ProDet~\citep{cheng2024prodet}
& NeurIPS'24
& 92.9 & 77.1 & 74.7 & 92.8 & 83.7 & 97.5 & 84.4 & 90.8 & 86.7 \\

Effort~\citep{yan2025effort}
& ICML'25
& 87.3 & 98.3 & 92.6 & 95.7 & 93.6 & 95.3 & 92.6 & 96.2 & 94.0 \\

Forensics Adapter~\citep{cui2025forensicsadapter}
& CVPR'25
& 87.2 & 96.3 & 94.1 & 96.8 & 94.6 & 97.4 & 91.4 & 94.2 & 94.0 \\

TSRL~\citep{lei2026tsrl}
& CVPR'26
& 90.2 & 98.3 & 94.8 & 97.5 & 93.7 & 93.3 & 90.4 & 95.4 & 94.2 \\

Video Blending + StA~\citep{yan2025vbsta}
& CVPR'25
& 90.6 & 98.0 & 91.2 & 96.4 & 93.7 & 94.6 & 93.1 & 96.0 & 94.2 \\

DFD-HR~\citep{sun2026dfdhr}
& CVPR'26
& 95.6
& \textbf{99.3}
& \underline{97.0}
& \underline{98.4}
& \underline{98.1}
& \underline{98.6}
& \underline{97.0}
& \textbf{98.8}
& \underline{97.8} \\

Query-token forgery purification~\citep{wang2026qtfp}
& CVPR'26
& \underline{96.5}
& \underline{99.1}
& \textbf{97.7}
& \textbf{98.6}
& \textbf{98.7}
& \textbf{99.0}
& \textbf{97.7}
& \textbf{98.8}
& \textbf{98.3} \\

\midrule

\textbf{ConsistencyNet (ours)}
& ---
& \textbf{97.2}
& 97.1
& 94.3
& 96.4
& 93.1
& 95.9
& 93.7
& 96.6
& 95.5 \\

\bottomrule
\end{tabular}%
}

\vspace{2pt}

{\scriptsize
Blend: BlendFace; FD: FaceDancer; IS: InSwapper;
MS: MobileSwap; SS: SimSwap; UF: UniFace.}

\end{table}
}

\FloatBarrier
\paragraph{Cross-manipulation evaluation.}
Finally, we evaluate transfer to eight unseen face-swap generators
from DF40(Table~\ref{tab:df40swap}). ConsistencyNet achieves 95.5 average AUC across eight unseen face-swap
generators, with AUC above 93 on every method and the highest performance
on BlendFace (97.2). The gap to Query-token forgery purification decreases
from 5.5 AUC on Celeb-DFv2 (Table~\ref{tab:crossdataset}) to 2.8 AUC on
DF40 on average. This consistent performance across diverse unseen
generators supports the transferability of the learned consistency
representation beyond the observed training manipulations.
during 

\subsection{Analysis of Learned Consistency Representations}
\label{sec:analysis:representation}
\begin{figure}[!htbp]
    \centering
    \includegraphics[width=\columnwidth]
    {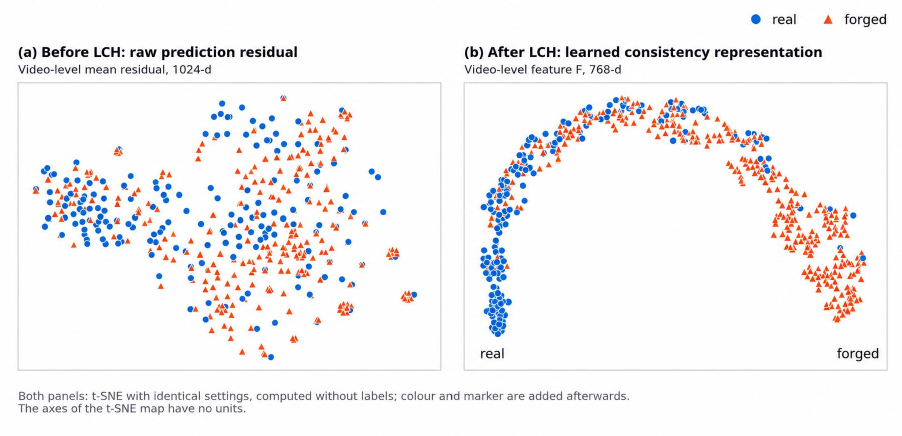}
    \caption{\textbf{Prediction residuals before and after LCH on unseen
    Celeb-DFv2 videos.}
    t-SNE visualization of (a) raw video-level prediction residuals and
    (b) the final consistency representations used for classification.
    Real and forged samples largely overlap before LCH, while the learned
    representations show clearer class separation after LCH.}
    \label{fig:representation_tsne}
\end{figure}

\paragraph{From prediction residuals to discriminative representations.}
To examine how LCH transforms the V-JEPA prediction residual, we visualize
video-level representations from the unseen Celeb-DFv2 test set using
t-SNE. Fig.~\ref{fig:representation_tsne} compares the raw prediction
residual before LCH with the final consistency representation used by the
classifier. The raw residuals show substantial overlap between authentic
and manipulated videos. After LCH, the representations become more
organized along a real-to-forged direction, with substantially greater
separation between the two classes. Importantly, Celeb-DFv2 is not used
during training, so this visualization illustrates how the learned
consistency representation transfers to an unseen dataset.

\subsection{Ablation Study}
\label{sec:exp:ablations}

\paragraph{Training objective.}
We first study how the consistency prior is imposed. Replacing the
pairwise margin with random groups (A1) reduces the average AUC by
2.5\%, despite improving Celeb-DF-v2 from 90.5 to 91.2.
Applying the prior only to class means (A2) gives a larger drop of
3.0\%. Reversing the margin to enforce higher residuals for real
clips than forged clips (A3) reduces the average by 2.2\%.
These results support applying the prior at the pair level with forged
clips above real clips.

\paragraph{Architecture.}
We next examine the role of the residual and predictor. Removing the
margin loss (B1) reduces the average AUC by 4.3\%, showing that
the consistency constraint is important for cross-dataset transfer.
Without the margin, using predicted features directly (B2) performs
better than using residuals, but remains 2.0\% below the full
model. Freezing the predictor (B3) causes the largest degradation,
with the average AUC dropping from 89.0 to 76.6. This indicates that
adapting the predictor together with the residual-based objective is
important for generalization.

\paragraph{Tail fraction.}
Finally, we vary the fraction of largest residual norms used to compute
the clip-level consistency score. The full model uses the largest
12.5\%. Reducing this fraction to 6.25\% (C1) lowers the average AUC
by 2.9\%, while increasing it to 25\% (C2) lowers AUC by
2.6\%. Thus, focusing on the largest 12.5\% provides the best
average performance among the tested settings.

\paragraph{Margin.}
The full model uses $m=2$ and $\lambda=0.1$. Reducing the margin to
$m=1$ results in a 2.4\% AUC drop, while increasing it to $m=4$ or
increasing the margin weight to $\lambda=0.2$ results in a 2.2\%
drop. The default setting therefore provides the strongest
cross-dataset performance.

\paragraph{Prediction lengths.}
The full model uses prediction lengths $\delta\in\{2,4\}$. Using only
$\delta=2$ results in a 2.0\% AUC drop, while using only $\delta=4$
reduces AUC by 2.7\%. Adding the longer prediction length
$\delta=6$ results in the largest drop of 3.0\%. This suggests that
combining $\delta=2$ and $\delta=4$ provides a better balance between
available context and prediction length.

{
\begin{table}[!htbp]
\centering
\small
\caption{\textbf{Ablation study.}
Video-level AUC (\%) on four unseen datasets. Avg. is the mean across
datasets and $\Delta$ is the change from the full model.}
\label{tab:ablation}

\setlength{\tabcolsep}{4.5pt}
\renewcommand{\arraystretch}{1.05}

\resizebox{\textwidth}{!}{%
\begin{tabular}{@{}clcccccc@{}}
\toprule
\textbf{\#}
& \textbf{Variant}
& \textbf{CDF-v2}
& \textbf{DFDC}
& \textbf{DFDCP}
& \textbf{WildDF}
& \textbf{Avg.}
& $\boldsymbol{\Delta}$ \\
\midrule

---
& \textbf{ConsistencyNet}
& 90.5 & 87.1 & 90.4 & 88.2
& \textbf{89.0} & --- \\

\midrule
\multicolumn{8}{l}{\textit{A. Training objective}} \\

A1
& No prior: random groups
& \textbf{91.2} & 82.1 & 89.1 & 83.7
& 86.5 & $-2.5$ \\

A2
& Prior on class means
& 86.0 & 85.2 & 87.7 & 84.9
& 86.0 & $-3.0$ \\

A3
& Reversed prior: real above forged
& 88.8 & 84.7 & 88.5 & 85.2
& 86.8 & $-2.2$ \\

\midrule
\multicolumn{8}{l}{\textit{B. Architecture}} \\

B1
& No margin ($\lambda=0$)
& 81.6 & 82.7 & \textbf{90.7} & 83.7
& 84.7 & $-4.3$ \\

B2
& Prediction instead of residual
& 86.5 & 85.4 & 88.8 & 87.4
& 87.0 & $-2.0$ \\

B3
& Frozen predictor
& 81.5 & 75.0 & 81.2 & 68.6
& 76.6 & $-12.4$ \\

\midrule
\multicolumn{8}{l}{\textit{C. Tail fraction}} \\

C1
& Largest 6.25\%
& 86.2 & 83.5 & 88.5 & 86.2
& 86.1 & $-2.9$ \\

C2
& Largest 25\%
& 88.4 & 85.8 & 84.8 & 86.4
& 86.4 & $-2.6$ \\

\midrule
\multicolumn{8}{l}{\textit{D. Margin}} \\

D1
& Margin $m=1$
& 87.0 & 84.3 & 88.5 & 86.7
& 86.6 & $-2.4$ \\

D2
& Margin $m=4$
& 88.7 & 84.9 & 87.1 & 86.5
& 86.8 & $-2.2$ \\

D3
& Margin weight $\lambda=0.2$
& 87.0 & 84.0 & 88.8 & 87.5
& 86.8 & $-2.2$ \\

\midrule
\multicolumn{8}{l}{\textit{E. Prediction lengths}} \\

E1
& One length, $\delta\in\{2\}$
& 88.5 & 84.0 & 89.0 & 86.5
& 87.0 & $-2.0$ \\

E2
& One length, $\delta\in\{4\}$
& 86.5 & 84.0 & 88.5 & 86.0
& 86.3 & $-2.7$ \\

E3
& Three lengths, $\delta\in\{2,4,6\}$
& 86.5 & 83.5 & 88.0 & 86.0
& 86.0 & $-3.0$ \\

\bottomrule
\end{tabular}%
}
\end{table}

\FloatBarrier

\section{Discussion and Limitations}
\label{sec:discussion}

Our results suggest that prediction residuals from a self-supervised video predictor provide a useful signal for cross-dataset deepfake detection. The ablations and matched sign-flip experiments further show how these residuals are used during training matters for generalization. Overall, the results demonstrate strong and consistent performance across diverse datasets and evaluation settings, supporting the ability of prediction-based consistency cues to generalize across manipulation domains.

\textbf{Limitations.}
Our system is visual-only and operates on face-cropped videos. We do
not evaluate audio-visual consistency, adversarial robustness, or
demographic fairness. The method also requires a predictive video
backbone, which limits direct comparison with encoder-only foundation
models. Future work could extend this study to other predictive video backbones, emerging deepfake benchmarks, and more comprehensive evaluations of robustness and fairness.

\subsection*{AI Use Statement}

We used Large Language Models (LLMs) solely to assist with language polishing and refinement during the preparation of this manuscript. Their use was limited to improving grammar, sentence structure, and readability. All research ideas, methodology, experiments, data analysis, results, and conclusions are the original work of the authors. The authors take full responsibility for the accuracy and integrity of the content presented in this paper.

\bibliographystyle{iclr2027_conference}
\bibliography{iclr2027_conference}

\appendix
\section{Appendix}

\section{Extended Related Work}
\label{app:extended_related_work}

\paragraph{Recent approaches to cross-dataset deepfake detection.}
Recent work addresses cross-dataset generalization from several
complementary directions. D3~\citep{d3_cvpr2025} uses multiple
generators together with a discrepancy-learning branch.
DiffusionFake~\citep{diffusionfake_neurips2024} incorporates features
from a frozen diffusion model to improve source--target
disentanglement. FreqBlender~\citep{freqblender_neurips2024}
constructs pseudo-forgeries through frequency-domain blending.
Video-level blending with a spatiotemporal adapter
(Plug-and-Play VLB)~\citep{plugnplay_vlb_cvpr2025} targets video
generalization through augmentation and parameter-efficient
adaptation. DFD-FCG~\citep{dfd_fcg_cvpr2025} adapts a pretrained
image encoder using facial-component-guided features.

ConsistencyNet approaches the problem from a different perspective.
Rather than introducing an external diffusion model, synthesizing
additional pseudo-forgeries, or relying primarily on frequency-domain
cues, we use the latent prediction residual of a pretrained
self-supervised video model. Specifically, the detector compares the
V-JEPA predictor output, after dimensional alignment through the
Bridge, with the corresponding stop-gradient encoder target.

\paragraph{Consistency-based forensic signals.}
RealForensics~\citep{realforensics} is related in its use of
self-supervised consistency for forgery detection. It learns
audio--visual correspondence as a consistency signal, whereas
ConsistencyNet is visual-only and derives its signal from disagreement
between predicted and target latent video representations. These
approaches therefore exploit different forms of consistency and are
complementary rather than directly equivalent.

\section{Visualization of Prediction Residuals}
\label{app:residual_visualization}

\paragraph{Real and fake prediction residuals.}
Figure~\ref{fig:residual_real_fake} visualizes the prediction residual
for an example real clip and an example fake clip. The prediction-error
maps show the spatial distribution of the residual magnitude for each
clip. The final panel shows the difference between the two maps,
computed as forged minus real. Positive values therefore indicate
regions where the prediction error is larger for the forged clip.

The example illustrates the consistency-margin prior introduced in
Section~\ref{sec:method:sfip}: the forged clip exhibits larger
prediction errors in several facial regions, producing positive values
in the forged-minus-real difference map. This visualization is
illustrative and is not intended as a quantitative estimate of the
margin across the dataset.

\begin{figure*}[t]
    \centering
    \includegraphics[
        width=\textwidth
    ]{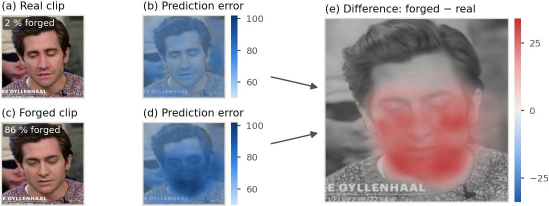}
    \caption{\textbf{Visualization of prediction residuals for real and
    forged clips.}
    (a) Real clip and (b) its prediction-error map; (c) forged clip and
    (d) its prediction-error map. (e) Difference between the forged and
    real prediction-error maps, computed as forged minus real. Positive
    values indicate regions with larger prediction error for the forged
    clip.}
    \label{fig:residual_real_fake}
\end{figure*}

\section{Implementation Details}
\label{app:method_details}

This section provides additional implementation details for
ConsistencyNet. Unless stated otherwise, the settings below are shared
across all experiments. The V-JEPA~2.1 encoder remains frozen. Within
the predictor, the transformer base weights remain frozen, while the
LoRA adapters and selected input/output components are trainable. The
Bridge, Latent Consistency Head (LCH), and classifier are also trained.

\paragraph{Input clips and latent representation.}
Each input consists of 16 face-aligned RGB frames at
$384\times384$ resolution. V-JEPA~2.1 maps the clip to
$T'=8$ temporal feature positions, with $S=576$ spatial tokens at
each position. The encoder therefore produces
\begin{equation}
\mathbf{H}
\in
\mathbb{R}^{B\times T'\times S\times1024},
\qquad
T'=8,\quad S=576,
\end{equation}
where $B$ denotes the batch size.

We use prediction lengths
$\mathcal{D}=\{2,4\}$. For prediction length $\delta$, the first
$T'-\delta$ temporal positions provide the predictor context and the
final $\delta$ positions define the prediction target. Thus, each
prediction length contains $\delta S$ target tokens: 1152 tokens for
$\delta=2$ and 2304 tokens for $\delta=4$.

\paragraph{Frozen encoder and predictor adaptation.}
The V-JEPA~2.1 encoder contains 24 transformer blocks and produces
1024-dimensional latent features. All encoder parameters remain frozen
during training.

The V-JEPA predictor contains 12 transformer blocks and produces
1664-dimensional target predictions. Its transformer base weights
remain frozen. We apply rank-16 LoRA to the query--key--value and
output projections of every predictor block. The predictor input
embedding, output projection, final layer normalization, and mask token
are additionally trainable.

The predictor LoRA modules contribute 442,368 trainable parameters.

\paragraph{Low-rank Bridge.}
The predictor and encoder operate at different feature dimensions:
the predictor output is 1664-dimensional, whereas the encoder target
is 1024-dimensional. We therefore map the predictor output into the
encoder feature space using a two-layer low-rank Bridge,
\begin{equation}
\mathcal{B}=B_2\circ B_1,
\end{equation}
with
\begin{equation}
B_1:
\mathbb{R}^{1664}\rightarrow\mathbb{R}^{64},
\qquad
B_2:
\mathbb{R}^{64}\rightarrow\mathbb{R}^{1024}.
\end{equation}

Both linear layers include biases. The Bridge therefore contains
\begin{equation}
(1664\times64+64)+(64\times1024+1024)
=
173{,}120
\end{equation}
trainable parameters.

The bottleneck is substantially smaller than a direct
$1664\rightarrow1024$ projection, which would require
$1.70$\,M parameters including bias.

\paragraph{Prediction residual construction.}
For each prediction length, we subtract the corresponding encoder
target from the bridged predictor output,
\begin{equation}
\mathbf{r}^{(\delta)}
=
\mathcal{B}
\left(
\mathbf{h}_{P}^{(\delta)}
\right)
-
\sg
\left(
\mathbf{h}_{E}^{(\delta)}
\right),
\end{equation}
where $\sg(\cdot)$ denotes stop-gradient.

The resulting residual tensor has shape
\begin{equation}
\mathbf{r}^{(\delta)}
\in
\mathbb{R}^{B\times\delta S\times1024}.
\end{equation}
Hence,
\begin{equation}
\mathbf{r}^{(2)}
\in
\mathbb{R}^{B\times1152\times1024},
\qquad
\mathbf{r}^{(4)}
\in
\mathbb{R}^{B\times2304\times1024}.
\end{equation}

The stop-gradient operation prevents the encoder target from being
updated by the detector losses. Gradients therefore update the
trainable predictor components, Bridge, and LCH while the pretrained
encoder remains fixed.

\paragraph{Predictor Attention-conditioned Stop-gradient Centering.}
Raw prediction residuals may contain systematic offsets associated
with how strongly individual target tokens depend on visible context.
PASC uses predictor attention to define two context-dependence groups
and centers their residuals separately.

We extract self-attention from predictor block~6. Let
$A_{i,j}^{(6,h)}$ denote the attention from target token $i$ to token
$j$ in head $h$. We average over the $H_a$ attention heads,
\begin{equation}
A_{i,j}^{(6)}
=
\frac{1}{H_a}
\sum_{h=1}^{H_a}
A_{i,j}^{(6,h)}.
\end{equation}

For target token $i$, its context-attention ratio is
\begin{equation}
c_i
=
\frac{
    \sum_{j\in\mathrm{ctx}}
    A_{i,j}^{(6)}
}{
    \sum_j A_{i,j}^{(6)}
}.
\end{equation}

Because the attention weights are post-softmax,
$c_i$ measures the fraction of the token's attention mass assigned
to visible context tokens.

Within each clip, we compute the median context-attention ratio and
define
\begin{equation}
\mathcal{H}
=
\left\{
i:c_i>\operatorname{median}(c)
\right\},
\qquad
\mathcal{L}
=
\left\{
i:c_i\leq\operatorname{median}(c)
\right\}.
\end{equation}

The two groups correspond to relatively high- and low-context target
tokens within the same clip. No real/fake label is used to construct
these groups.

For each residual token, PASC subtracts the mean residual of its
assigned group:
\begin{equation}
\tilde{\mathbf{r}}_i^{(\delta)}
=
\mathbf{r}_i^{(\delta)}
-
\boldsymbol{\mu}_{\mathcal{G}(i)}^{(\delta)},
\end{equation}
where
\begin{equation}
\boldsymbol{\mu}_{g}^{(\delta)}
=
\frac{1}{|g|}
\sum_{j\in g}
\mathbf{r}_{j}^{(\delta)},
\qquad
g\in\{\mathcal{H},\mathcal{L}\}.
\end{equation}

The attention values used to determine group membership are detached
from the gradient. The centering operation itself remains
differentiable with respect to the residual features. PASC therefore
introduces no trainable parameters.

\paragraph{Training and inference centering.}
PASC requires predictor attention only during training. At inference
time, the attention hook is removed. We reshape the residual tokens
according to their predicted temporal positions and subtract one
spatial mean independently from each predicted frame.

For temporal position $t$,
\begin{equation}
\boldsymbol{\mu}_{t}^{(\delta)}
=
\frac{1}{S}
\sum_{s=1}^{S}
\mathbf{r}_{t,s}^{(\delta)},
\end{equation}
and inference-time centering becomes
\begin{equation}
\tilde{\mathbf{r}}_{t,s}^{(\delta)}
=
\mathbf{r}_{t,s}^{(\delta)}
-
\boldsymbol{\mu}_{t}^{(\delta)}.
\end{equation}

Thus, predictor attention is not required during deployment while the
remaining LCH operations are unchanged.

\paragraph{Magnitude--direction decomposition.}
After centering, each residual token is separated into magnitude and
direction. The normalized direction is
\begin{equation}
\hat{\mathbf{r}}_i^{(\delta)}
=
\frac{
    \tilde{\mathbf{r}}_i^{(\delta)}
}{
    \max\left(
        \|\tilde{\mathbf{r}}_i^{(\delta)}\|_2,
        \sqrt{\varepsilon}
    \right)
},
\qquad
\varepsilon=10^{-12}.
\end{equation}

The normalized direction is provided to LCH for classification. Raw
residual magnitude is used separately by the consistency-margin
objective described in Section~\ref{sec:method:training}; it is not
concatenated with the classification representation.

\paragraph{Prediction-length-specific projection heads.}
For each prediction length
$\delta\in\{2,4\}$, LCH uses $K=4$ parallel projection heads. Head $k$
computes
\begin{equation}
\mathbf{e}_{i}^{(\delta,k)}
=
\operatorname{LN}
\left(
W_{\delta}^{(k)}
\hat{\mathbf{r}}_{i}^{(\delta)}
+
\mathbf{b}_{\delta}^{(k)}
\right),
\end{equation}
where
\begin{equation}
W_{\delta}^{(k)}
\in
\mathbb{R}^{768\times1024},
\qquad
k=1,\ldots,4.
\end{equation}

Each projection therefore produces a 768-dimensional representation
while preserving the token structure. The four heads are instantiated
independently for each prediction length.

\paragraph{Projection-head gating.}
The four projections are combined using a learned softmax gate rather
than a uniform average. We use one trainable scalar logit per
projection head for each prediction length:
\begin{equation}
\boldsymbol{\gamma}_{\delta}
=
\operatorname{softmax}
\left(
\boldsymbol{a}_{\delta}
\right),
\qquad
\boldsymbol{a}_{\delta}\in\mathbb{R}^{4}.
\end{equation}

The resulting token representation is
\begin{equation}
\mathbf{e}_{i}^{(\delta)}
=
\sum_{k=1}^{4}
\gamma_{\delta,k}
\mathbf{e}_{i}^{(\delta,k)}
\in
\mathbb{R}^{768}.
\end{equation}

This gate contributes
$4\times2=8$ trainable parameters.

\paragraph{Spatial Anomaly Pooling.}
For prediction length $\delta$, the projected representation contains
$\delta S$ spatial tokens:
\begin{equation}
\mathbf{E}^{(\delta)}
\in
\mathbb{R}^{B\times\delta S\times768}.
\end{equation}

We reshape it as
\begin{equation}
\mathbf{E}^{(\delta)}
\rightarrow
\mathbb{R}^{B\times\delta\times S\times768}.
\end{equation}

SAP first normalizes the spatial-token features within each sample.
A learned scalar scoring layer then assigns a score to each spatial
token:
\begin{equation}
q_{t,s}^{(\delta)}
=
\mathbf{w}_{\rm SAP}^{\top}
\mathbf{e}_{t,s}^{(\delta)}
+
b_{\rm SAP},
\end{equation}
with
$\mathbf{w}_{\rm SAP}\in\mathbb{R}^{768}$.
The spatial weights are obtained by
\begin{equation}
\alpha_{t,s}^{(\delta)}
=
\operatorname{softmax}_{s}
\left(
q_{t,s}^{(\delta)}
\right).
\end{equation}

The pooled representation is
\begin{equation}
\mathbf{p}_{t}^{(\delta)}
=
\sum_{s=1}^{S}
\alpha_{t,s}^{(\delta)}
\mathbf{e}_{t,s}^{(\delta)}.
\end{equation}

A separate SAP scoring layer is used for each prediction length, giving
\begin{equation}
2(768+1)=1{,}538
\end{equation}
trainable SAP parameters.

SAP therefore maps
\begin{equation}
[B,\delta,S,768]
\longrightarrow
[B,\delta,768].
\end{equation}

\paragraph{Temporal aggregation.}
For each prediction length, SAP outputs one representation for each
predicted temporal position. We average these representations:
\begin{equation}
\mathbf{F}_{\delta}
=
\frac{1}{\delta}
\sum_{t=1}^{\delta}
\mathbf{p}_{t}^{(\delta)}
\in
\mathbb{R}^{768}.
\end{equation}

\paragraph{Cross-length aggregation.}
We combine the two prediction-length representations using a learned
softmax gate. One scalar logit is assigned to each prediction length,
\begin{equation}
\boldsymbol{\beta}
=
\operatorname{softmax}
\left(
\mathbf{b}_{\rm len}
\right),
\qquad
\mathbf{b}_{\rm len}\in\mathbb{R}^{2}.
\end{equation}

The final consistency representation is
\begin{equation}
\mathbf{F}
=
\beta_{2}\mathbf{F}_{2}
+
\beta_{4}\mathbf{F}_{4}
\in
\mathbb{R}^{768}.
\end{equation}

The cross-length gate contributes two trainable parameters.

\paragraph{Classifier.}
The final classifier is a two-layer MLP,
\begin{equation}
768\rightarrow384\rightarrow2.
\end{equation}

Given the aggregated consistency representation $\mathbf{F}$, it
produces
\begin{equation}
\mathbf{z}
=
\operatorname{MLP}(\mathbf{F})
\in
\mathbb{R}^{2},
\end{equation}
corresponding to the real and fake logits.

\paragraph{Gradient flow.}
The encoder target is detached when the prediction residual is
constructed, and all pretrained encoder parameters remain frozen.
Gradients from the classification objective propagate through the
classifier, cross-length aggregation, temporal aggregation, SAP,
projection heads, Bridge, and trainable predictor components.
The consistency-margin objective acts on raw residual magnitude and
propagates through the Bridge and predictor trainable components while
leaving the stop-gradient encoder target unchanged.

This separates the pretrained target representation from the
task-specific adaptation while allowing the lightweight trainable
modules to shape the prediction residual for forensic detection.

\paragraph{End-to-end tensor shapes.}
For clarity, the principal transformations for one clip are
\begin{align}
\text{encoder features}
&:
[B,8,576,1024],
\\
\text{predictor target features}
&:
[B,\delta\!\times\!576,1664],
\\
\text{bridged prediction}
&:
[B,\delta\!\times\!576,1024],
\\
\text{prediction residual}
&:
[B,\delta\!\times\!576,1024],
\\
\text{parallel projections}
&:
[B,4,\delta\!\times\!576,768],
\\
\text{head-aggregated features}
&:
[B,\delta\!\times\!576,768],
\\
\text{SAP output}
&:
[B,\delta,768],
\\
\text{temporal aggregation}
&:
[B,768],
\\
\text{cross-length aggregation}
&:
[B,768],
\\
\text{classifier output}
&:
[B,2].
\end{align}

For $\delta=2$, the residual contains 1152 target tokens; for
$\delta=4$, it contains 2304 target tokens.

\paragraph{Parameter accounting.}
Table~\ref{tab:parameter_accounting} reports the parameter count of each
component in the instantiated architecture. LCH uses four parallel
projection heads for each prediction length. The learned
projection-head, SAP, and cross-length gates are included explicitly
in the trainable parameter count.

\begin{table*}[t]
\centering
\scriptsize
\caption{\textbf{Parameter accounting for ConsistencyNet.}
Trainable and frozen parameters are counted from the stated
architecture. LCH uses four parallel projection heads for each
prediction length, together with learned projection-head, SAP, and
cross-length gates.}
\label{tab:parameter_accounting}

\setlength{\tabcolsep}{3pt}
\renewcommand{\arraystretch}{1.08}

\begin{tabularx}{\textwidth}{@{}
p{0.16\textwidth}
X
r
r
@{}}
\toprule
\textbf{Component}
& \textbf{Layer}
& \textbf{Trainable}
& \textbf{Frozen} \\
\midrule

Encoder
& V-JEPA~2.1 ViT-L (24 blocks)
& ---
& 304,680,960 \\

\midrule

\multirow{5}{*}{Predictor}
& Base weights (12 blocks)
& ---
& 21,937,664 \\

& Predictor LoRA on query--key--value and output projections
  ($A$: 147,456; $B$: 294,912)
& 442,368
& --- \\

& Input embedding, $1024\rightarrow384$
& 393,600
& --- \\

& Output projection, $384\rightarrow1664$
& 640,640
& --- \\

& Final layer normalization and mask token
& 1,152
& --- \\

\cmidrule(lr){2-4}
& \textit{Predictor total}
& \textit{1,477,760}
& \textit{21,937,664} \\

\midrule

Bridge $\mathcal{B}$
& $1664\rightarrow64\rightarrow1024$
& 173,120
& --- \\

\midrule

\multirow{6}{*}{LCH}
& PASC / context-conditioned centering
& 0
& --- \\

& $\ell_2$ residual normalization
& 0
& --- \\

& Four parallel projection heads per prediction length,
  $1024\rightarrow768$, with LayerNorm
& 6,309,888
& --- \\

& Learned projection-head gate
& 8
& --- \\

& Spatial Anomaly Pooling, including learned spatial gate
& 1,538
& --- \\

& Temporal aggregation
& 0
& --- \\

\cmidrule(lr){2-4}
& \textit{LCH total, two prediction lengths}
& \textit{6,311,434}
& --- \\

\midrule

Cross-length aggregation
& Learned two-way softmax gate
& 2
& --- \\

\midrule

Classifier
& MLP $768\rightarrow384\rightarrow2$
& 296,066
& --- \\

\midrule

\textbf{Total}
& \textbf{334,877,006 parameters}
& \textbf{8,258,382 (2.47\%)}
& \textbf{326,618,624} \\

\bottomrule
\end{tabularx}
\end{table*}

\section{From V-JEPA Prediction to the Consistency-Margin Prior}
\label{app:jepa_to_prior}

V-JEPA learns to predict masked latent representations from visible
spatio-temporal context. Abstractly, its self-supervised prediction
objective can be written as
\begin{equation}
\mathcal{L}_{\mathrm{JEPA}}(v)
=
\mathbb{E}_{m\sim\mathcal{M}}
\left[
\left\|
P_{\psi}\!\left(
E_{\phi}(v\odot m),m
\right)
-
\sg\!\left(
E_{\phi}(v)
\right)_{m}
\right\|_2^2
\right],
\label{eq:jepa_loss}
\end{equation}
where $E_{\phi}$ denotes the encoder, $P_{\psi}$ the predictor,
$\mathcal{M}$ the masking distribution, $m$ a sampled mask, and
$\sg(\cdot)$ the stop-gradient operator. The predictor is therefore
trained to estimate latent target representations from visible
context rather than reconstructing pixels directly.

ConsistencyNet uses the discrepancy between the predicted and target
latent representations as a forensic signal. For prediction length
$\delta$, we compute
\begin{equation}
\mathbf{r}^{(\delta)}
=
\mathcal{B}\!\left(
\mathbf{h}_{P}^{(\delta)}
\right)
-
\sg\!\left(
\mathbf{h}_{E}^{(\delta)}
\right),
\end{equation}
where $\mathcal{B}$ aligns the predictor representation with the
encoder feature space. The residual therefore measures the agreement
between a target representation predicted from visible context and
the corresponding encoder representation.

\paragraph{Consistency-margin prior.}
As introduced in Section~\ref{sec:intro}, we hypothesize that fake
videos produce larger prediction residuals than real videos on average,
$\mathbb{E}[\|\mathbf{r}_{\rm fake}\|_2] >
\mathbb{E}[\|\mathbf{r}_{\rm real}\|_2]$.
This relation is used as an empirical training prior rather than a
guaranteed property of V-JEPA. Residual magnitude can also depend on
video content, compression, sampling, manipulation type, and dataset
shift. We therefore do not require the ordering to hold for every
individual clip.

This distinction motivates the Sign-Flip Identification Protocol
(SFIP) in Section~\ref{sec:method:sfip}. SFIP reverses the direction
of the margin while keeping the remaining training setup fixed and
matches the two settings on FF++ validation AUC before cross-dataset
evaluation. Across the evaluated cross-dataset benchmarks, the
reversed-margin setting exhibits a larger generalization gap on
average, providing empirical evidence that the direction of the
consistency margin is relevant to cross-dataset transfer in our
training setup.

\section{Training and Evaluation Details}
\label{app:hparams}

\paragraph{Training and in-domain evaluation.}
We train ConsistencyNet only on FaceForensics++ (FF++).
Each training sample consists of a 16-frame face clip. Faces are detected and aligned before being processed by the
V-JEPA input pipeline. FF++ validation is used for in-domain
evaluation and to match the standard and reversed margin settings
in SFIP.

\paragraph{Cross-dataset evaluation.}
We evaluate the FF++-trained model without further fine-tuning or
adaptation on Celeb-DFv2~\citep{li2020celebdf}, DFDC~\citep{dolhansky2020dfdc},
DFDCP, WildDeepfake~\citep{zi2020wilddeepfake}, and
Deepfake-Eval-2024. No samples or labels from these datasets are used
during training. We report video-level AUC (\%), obtained by
aggregating predictions from 16-frame clips into a video-level score.

\paragraph{Inference protocol.}
At inference time, predictor-attention extraction for PASC is disabled,
and the prediction residual is centered using a per-frame spatial
mean. All remaining LCH operations are unchanged. Clip-level
predictions are aggregated to obtain the final video-level score.

\paragraph{Training hyperparameters.}
Table~\ref{tab:training_hparams} summarizes the model, objective,
optimization, data, and compute settings used in the main experiments.

\begin{table*}[t]
\centering
\scriptsize
\caption{\textbf{Training hyperparameters of ConsistencyNet.}
The same configuration is used across all reported runs.}
\label{tab:training_hparams}

\setlength{\tabcolsep}{3.5pt}
\renewcommand{\arraystretch}{1.10}

\begin{tabular*}{\textwidth}{@{\extracolsep{\fill}}
p{0.10\textwidth}
p{0.25\textwidth}
p{0.60\textwidth}
@{}}
\toprule
\textbf{Group}
& \textbf{Hyperparameter}
& \textbf{Value} \\
\midrule

\multirow{8}{*}{\textit{Model}}
& Backbone
& V-JEPA~2.1 ViT-L/16~\citep{vjepa2_1_2025}, distilled from ViT-G;
  encoder frozen \\

& Input
& 16-frame face clip, $384\times384$; tubelet size 2, yielding
  $T'=8$ temporal positions with 576 spatial tokens per position \\

& Prediction lengths $\delta$
& $\{2,4\}$ \\

& Predictor adaptation
& LoRA on the query--key--value and output projections of all
  12 predictor blocks; rank $r=16$, $\alpha=16$ \\

& LoRA initialization
& $A$: Kaiming-uniform;
  $B\sim\mathcal{N}(0,10^{-6})$ \\

& Predictor layers trained in full
& Input embedding, output projection, final layer normalization,
  and mask token \\

& Bridge $\mathcal{B}$
& Linear $1664\rightarrow64\rightarrow1024$ \\

& Classifier
& MLP $768\rightarrow384\rightarrow2$ \\

\midrule

\multirow{5}{*}{\textit{Objective}}
& Loss
& Class-weighted cross-entropy
  $+\lambda\mathcal{L}_{\mathrm{margin}}$ \\

& Cross-entropy class weights
& Real : fake $=3:1$ \\

& Label smoothing
& $0.1$ \\

& Margin
& Pairwise; $m=2.0$, $\lambda=0.1$ \\

& Tail fraction of $n$
& $12.5\%$ \\

\midrule

\multirow{8}{*}{\textit{Optimization}}
& Optimizer
& Adam; $\beta=(0.9,0.999)$, $\epsilon=10^{-8}$ \\

& Learning rate
& $1\times10^{-4}$; 5 warm-up epochs followed by cosine decay
  to $10^{-6}$ \\

& Weight decay
& $5\times10^{-4}$ \\

& Gradient clipping
& Global norm $1.0$ \\

& Epochs / steps
& $50$ / $4{,}600$ \\

& Batch
& 384 clips ($16$ GPUs $\times$ 24 clips/GPU) \\

& Precision
& BF16 mixed precision \\

& Checkpoint
& Final epoch; no checkpoint selection \\

\midrule

\multirow{3}{*}{\textit{Data}}
& Training set
& FaceForensics++ \\

& Augmentation
& Clip-consistent augmentation: random crop with scale
  $[0.85,1.0]$ and center shift $\pm10\%$;
  horizontal flip ($p=0.5$);
  affine transformation with scale $[0.85,1.15]$, rotation
  $\pm5^\circ$, translation $\pm5\%$, and shear $\pm3^\circ$
  ($p=0.5$);
  one of color, gamma, brightness--contrast, or
  hue--saturation jitter ($p=0.5$);
  illumination gradient ($p=0.5$);
  Gaussian or motion blur ($p=0.2$);
  JPEG compression with quality $[20,100]$ ($p=0.5$) \\

& Normalization
& ImageNet mean and standard deviation \\

\midrule

\multirow{2}{*}{\textit{Compute}}
& Hardware
& 2 nodes $\times$ 8 NVIDIA A100 80\,GB GPUs \\

& Training time
& 9.2 hours \\

\bottomrule
\end{tabular*}
\end{table*}

\paragraph{LoRA initialization.}
For each adapted predictor projection, LoRA computes
\begin{equation}
\mathbf{y}
=
W\mathbf{x}
+
\frac{\alpha}{r}BA\mathbf{x},
\label{eq:lora_update}
\end{equation}
where the pretrained weight $W$ remains frozen. We use rank $r=16$
and $\alpha=16$, giving $\alpha/r=1$.

Following the reference LoRA initialization~\citep{lora2022hu},
$A$ is initialized with Kaiming-uniform weights. We initialize
$B\sim\mathcal{N}(0,10^{-6})$ rather than setting $B$ exactly to zero.
With $B=0$, the initial LoRA update $BA$ is zero and the gradient with
respect to $A$ is also zero at the first optimization step;
consequently, only $B$ is updated initially. The choice of which LoRA
factor is initialized at zero can affect fine-tuning dynamics.

Our small non-zero initialization allows gradients to reach both
factors from the first optimization step while keeping the initial
LoRA perturbation negligible relative to the pretrained weights. For
the query--key--value projection of the first predictor block, the
initial product $BA$ has a standard deviation of approximately
$1.2\times10^{-7}$, compared with $6.0\times10^{-2}$ for the
corresponding frozen weights.

Non-zero initialization of both LoRA factors has also been reported
to reduce sensitivity to the learning rate.
We use this initialization consistently throughout our experiments.
We did not perform a controlled comparison against the standard
$B=0$ initialization; therefore, we treat it as an implementation
choice rather than an empirical contribution.

\end{document}